\documentclass[letterpaper, 10 pt, conference]{ieeeconf} 

\IEEEoverridecommandlockouts    

\usepackage{graphicx}
\usepackage{amsmath, amssymb}
\usepackage{algorithm2e, setspace}
\usepackage{pifont}
\newcommand{\xmark}{\ding{55}}%
\newcommand{\tmark}{$\sim$}
\newcommand{\cmark}{\checkmark}
\usepackage{tikz}
\usepackage{accents}
\usepackage{hyperref}

\usepackage{rotating}
\usepackage{multirow}
\usepackage{color}
\usepackage[parfill]{parskip}
\usepackage[symbol]{footmisc}

\usepackage{mathtools}
\DeclareMathOperator*{\argminB}{argmin} 

\newcommand{\Evan}[1]{{\leavevmode\color{blue}#1}}
\newcommand{\Max}[1]{{\leavevmode\color{blue}#1}}
\newcommand{\myvec}[1]{\accentset{\rightharpoonup}{{#1}}}

\newcommand\rwbf[1]{#1}

\title{\LARGE \bf FlowControl: Optical Flow Based Visual Servoing}

\author{Max Argus$^1$, Lukas Hermann$^1$, Jon Long$^2$, Thomas Brox$^1$
\thanks{$^{1}$Computer Science, University of Freiburg, Germany; $^2$Symbio Robotics, Emeryville, USA.\newline Corresponding author: {\tt\small argusm@cs.uni-freiburg.de}}
}

\begin{document}

\maketitle
\thispagestyle{empty}
\pagestyle{empty}

\begin{abstract}


One-shot imitation is the vision of robot programming from a single demonstration, rather than by tedious construction of computer code.
We present a practical method for realizing one-shot imitation for manipulation tasks, exploiting modern learning-based optical flow to perform real-time visual servoing.
Our approach, which we call FlowControl, continuously tracks a demonstration video, using a specified foreground mask to attend to an object of interest.
Using RGB-D observations, FlowControl requires no 3D object models, and is easy to set up.
FlowControl inherits great robustness to visual appearance from decades of work in optical flow.
We exhibit FlowControl on a range of problems, including ones requiring very precise motions, and ones requiring the ability to generalize.

\end{abstract}

\section{Introduction}

The difficulty of robot programming is one of the central hurdles to the widespread  application of robots.
This task requires domain-specific expertise, making it inaccessible to untrained personnel and resulting in high system costs that lead to low adoption rates.
Few-shot imitation from videos is an appealing alternative to overcome this problem, as videos typically capture all task-relevant information.
%
However, the high dimensionality of videos makes it challenging to convert a demonstration video into actionable commands, while at the same time being robust to variations in the environment and the task.

The visual imitation problem can be divided into three separate components:
\vspace{-.8em}
\begin{enumerate}
    \item determining the salient objects relevant to the task,
    \item establishing correspondences between demonstration and the live application, and
    \item controlling the robot in order to reproduce the motion observed in the demonstration. 
\end{enumerate}
Each component is a difficult task.


Existing learning-based approaches need large amounts of training data. Other methods that rely on explicit pose estimation require precise 3D models of the objects. Even when given a 3D model, robust 6D pose estimation under appearance variation is an ongoing area of research.

In this paper, we propose a one-shot imitation learning\footnote{Ours is not a learning method in the sense of machine learning, we do not use data to optimize weights.}
approach which can robustly replicate a task from a single demonstration video, despite substantial variation of the objects' initial positions, orientations, and appearances.
Our approach imitates demonstrations through the use of learned optical flow;
point correspondences from optical flow together with a given foreground mask align live observations with demonstration frames.
After successfully aligning the live observation with the first demonstration frame, we successively do this for subsequent frames, thereby tracking an entire demonstration trajectory.
Thus, this formulation naturally extends to learning multi-step tasks. There is neither a need for CAD models of the objects involved, nor expensive pretraining in elaborate simulation environments.

While conceptually straightforward, our approach shows a large degree of robustness towards various factors of variation.
We successfully learn a variety of tasks, including picking and insertion of objects. The method is both data-efficient and achieves high success rates. 

The main contribution of our work is a practical, data-efficient approach to imitation which exploits and transfers the trained robustness of modern optical flow methods to robot control.

\begin{figure}[!t]
\centering
 \centering
 \includegraphics[width=\linewidth]{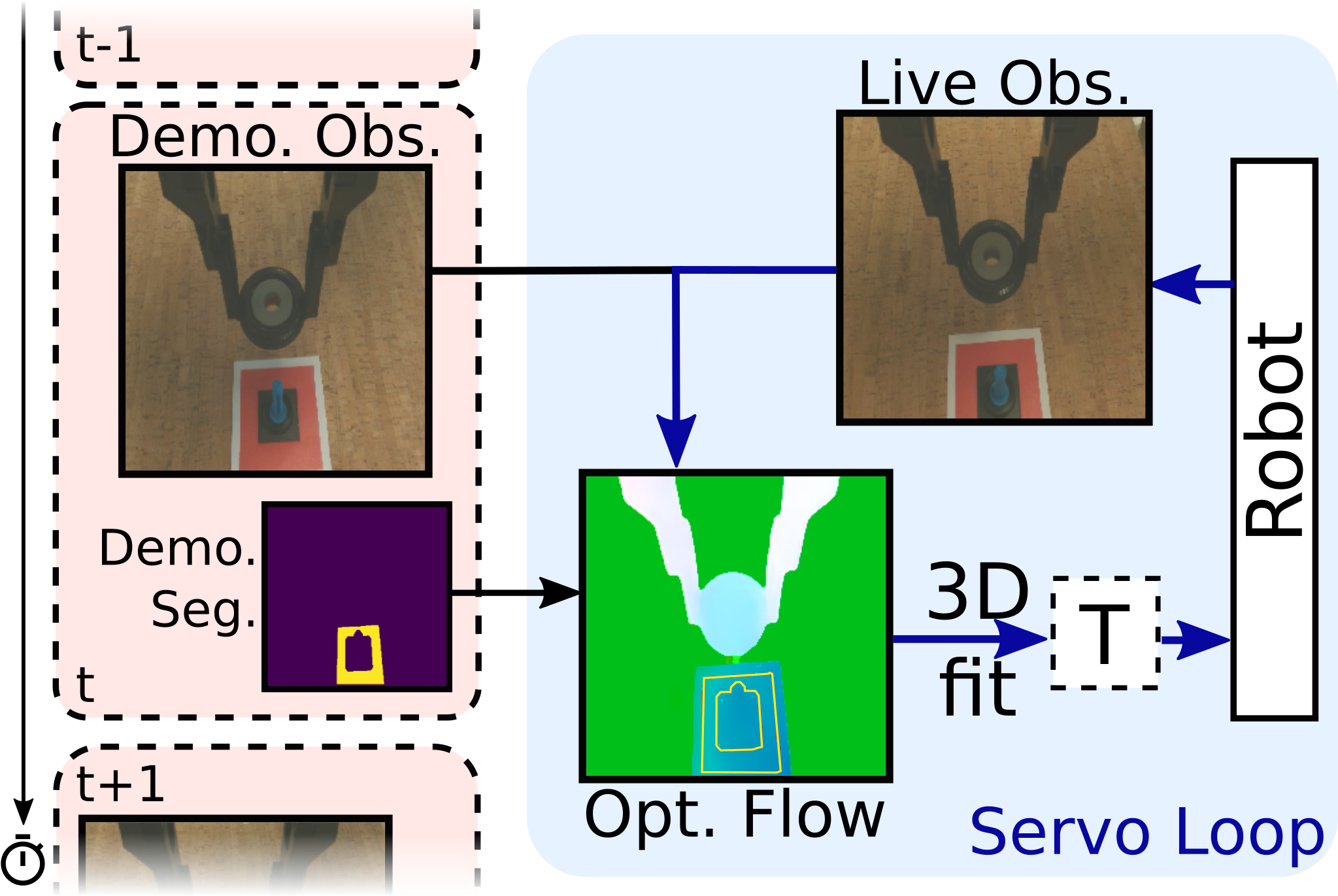}
 \caption{FlowControl follows demonstrations, shown in red, by using optical flow to find correspondences to a live camera image, shown in blue.
 This allows us to fit a rigid body transformation T, which aligns the state with the demonstration. The resulting servoing loop is shown with blue arrows.
 A foreground segmentation provided for the demonstration images (yellow mask) restricts correspondences to the object of interest.
 By repeating alignment for successive frames, we track trajectories so that tasks can be imitated. Figure \ref{fig:method} shows more details of the transformation computation.
 }
 \label{fig:teaser}
\end{figure}

\section{Related Work}


\rwbf{Imitation learning}\footnote{Also know as: Learning from Demonstration, Robot Programming by Demonstration, and Apprenticeship Learning} tries to control robots in such a way as to replicate what has been demonstrated. 
This approach reduces or eliminates the need for explicit programming~\cite{Billard08}.
It is often formulated as a learning problem \cite{Osa18} and numerous strategies to utilize demonstrations exist.
These include kinesthetic teaching, the decomposition of movement into motion primitives \cite{Kulic12}, additional exploration guidance for learning algorithms \cite{Aytar18}, and the direct learning of input to action mappings.

Some approaches start with low-dimensional inputs, e.g. \cite{Zimmermann18}, but most start from high-dimensional sensor input and aim to reduce its dimensionality.
This is often done by combining \rwbf{embeddings and imitation}.
Time Contrastive Networks \cite{Sermanet17} use multiple perspectives to learn a perspective-invariant embedding, and embeddings of images are used to guide Atari play in \cite{Aytar18}.
Another approach to the problem is one-shot imitation learning, in which policy networks are fine-tuned on conditioning demonstrations \cite{Yu18}.
In contrast, our approach need not learn an encoding to work directly with the high-dimensional sensor input.

\rwbf{Visual Servoing} (VS) is the concept of creating a feedback loop in which sensor data is used to compute control commands that change sensor measurements~\cite{Kragic02, Chaumette06}.
This feedback actively updates the control based on current observations, allowing for more adaptive control that is robust to variation in the environment \cite{Billard19}.
Many tasks in robotics, such as navigation, manipulation and learning from demonstration can be addressed using visual servoing \cite{Hutchinson96, Kragic02, Bandera12}.
Visual servoing allows the specification of goal configurations as target feature states to which a control law can servo.

Our approach is a form of visual servoing.
Visual servoing can be realized in a number of different ways.
It generally consists of three components: (1) image feature extraction, (2) the control law to decide where to move with respect to these features, and (3) joint control to execute this decision. 
While many works focus on formulating robust control laws and image feature extraction in the context of navigation, we consider the imitation of manipulation tasks.
Unlike most other visual servoing techniques ours benefits from the use of robust correspondences to generalize 
over scene geometry, lighting conditions, and object appearance.




Visual servoing is used in diverse applied robotics fields:
aircraft manufacturing \cite{Kheddar19}, 
robotic surgery \cite{Huang17},
marine ROVs \cite{Sivcev18}, and aerial manipulation \cite{Ramonsoria19}.
Recent methodological extensions range from
the combination with pose estimation \cite{Vakanski17} to
servoing to bounding boxes detected by an R-CNN \cite{Joffe18}.

A combination of \rwbf{visual servoing and optical flow} is often used to track poses  when servoing with respect to explicit pose estimates, which can be expensive to compute ab initio \cite{Nelson93, Nelson95, Vincze02}.
Optical flow was used for visual servoing by \cite{Mitsuda99}, where flow replaces template matching for visual servoing in an industrial positioning system. 
%
In \cite{Lopez19, Lopez19b}, a character navigation policy is learned based on the intermediate representation of optical flow.

A number of recent works combine \rwbf{visual servoing and learning}.
Some policy learning architectures bear a similarity to visual servoing, e.g. through the use of soft-argmax activations \cite{Levine15} or the use of optical flow as an auxiliary task \cite{Finn16}.
A number of applied works have also been published that use a combination of visual servoing and learning-based approaches:
%
\cite{Castelli17} trains a model acting as control law based on limited examples for electrical engine construction,
%
\cite{Sadeghi18} uses images as templates, and
%
\cite{Bateux18} trains a network to predict relative poses between images.
%
\cite{Lee19} implements target following by learning features and dynamics using reinforcement learning.
\cite{Triyonoputro19} presented learning-based visual servoing for peg insertion.

%
%

The performance of \rwbf{optical flow} computation has developed very rapidly in recent years \cite{Fischer15, Ilg16}. 
While initial interest in the optical flow problem was grounded in the context of active motion \cite{Gibson79}, it is also treated as an independent problem. 
Good performance of these methods has renewed interest in applications of optical flow \cite{Guney19}.
We benefit from the ability of recent learning-based optical flow to generalize, as it has been trained to be robust to common variations of appearance changes.

We use optical flow to solve the dense \rwbf{correspondence} problem. Optical flow is commonly defined as the per pixel apparent motion between two consecutive frames, which implies a data distribution. Strictly speaking, we apply an optical flow algorithm outside of its canonical scope.
This is not a trivial change since it adversely affects the data distribution; incidence of large displacements and out-of-frame occlusions increases.
Learned dense correspondences have also been generated in \cite{Bristow15, Choy16}. 
Similar to our method, \cite{Florence18, Manuelli19} learn to transfer key point affordances used for grasping.

\begin{figure*}[!t]
\centering
 \centering
 \includegraphics[width=.98\linewidth]{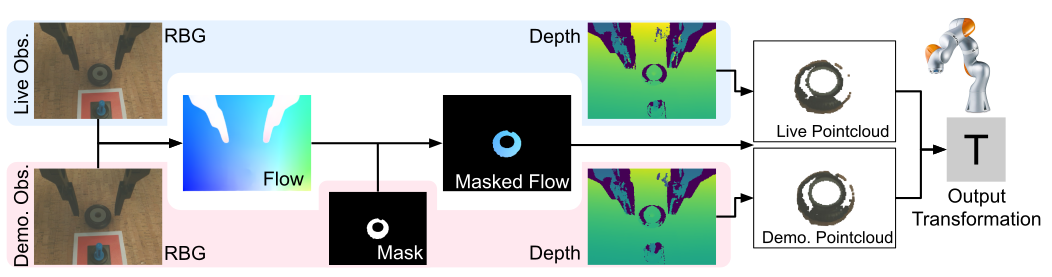}
 \caption{\textbf{FrameAlign} procedure to compute transformations from a live observation to a demonstration image. Optical flow is computed between the two images and masked with the given foreground mask. 
 The points undergo an inverse projection to 3D using the observed depth images. The resulting 3D point correspondences yield the rigid transformation necessary to better align the live state with the demonstration. Due to the large number of correspondences, the method is robust to noisy depth measurement and incomplete foreground masks.}
 \label{fig:method}
\end{figure*}

\section{FlowControl}

The main idea of FlowControl is to align a live video frame with a demonstration consisting of a sequence of target frames; this is illustrated in Figure \ref{fig:method}.
A learned optical flow algorithm is used to compute correspondences between a live frame and a target frame. Together with the recorded depth this yields a 3D flow of the underlying point cloud, which describes how points in the scene must move to reach the target state. A foreground mask restricts the points to a subset relevant for the task. 
These foreground points are used for computing the 3D transformation that brings the current frame closer to the target frame, and the robot is moved according to that transformation.

After successfully aligning with the first demonstration frame, the procedure can be successively repeated for subsequent frames. This tracks an entire demonstration trajectory, which can include the manipulation of multiple different objects. 

FlowControl benefits from three design choices: 
(1) the end-of-arm camera setup makes it easy to convert image transformations into the end-effector frame; 
(2) providing a first-person demonstration avoids the problem of a changing perspective;
(3) the learned optical flow is robust to many appearance variations. 

The foreground mask defines which object in the scene to align in order to progress to the next target state. While such a foreground mask could be inferred automatically from the demonstration video, we simplify the problem by manually providing the foreground mask, trading algorithmic complexity for an intuitive and practical extra step.
The foreground mask need not be precise; an under-segmentation of the object suffices. 


\textbf{Optical Flow:}
For computing correspondences, we use the optical flow from FlowNet2~\cite{Ilg16}. FlowNet2 is trained on FlyingThings3D~\cite{Mayer16}, a synthetic dataset procedurally assembled from a large set of different objects.
Data augmentation employed in FlowNet training ensures that the resulting optical flow is robust to changes in lighting and partial occlusions. 
It also performs well on textureless objects, having learned from a large set of different shapes.
While we benefit from the robustness and generalization this offers, correspondence is a modular component in our framework, and we could employ other methods such as \cite{Anderson16}, which does not require any learning.
Since the foreground mask is provided for the target image from the demonstration, we compute the optical flow from the target image to the live image. 


\textbf{Frame Alignment:}
We use pixel correspondences from the flow algorithm together with our aligned RGB-D data to match 3D points between the recorded demonstration observation and the live observations.
This gives us correspondences between individual points of the pointclouds.
Subsequently, we use SVD to compute a least-squares rigid transformation between these point clouds \cite{Horn54}.
This is an estimate of the relative transformation between the demonstration scene and the scene as observed live.
Servoing in this direction will align the camera image with the demonstration image.
We use a position-based controller to convert this transformation into a control signal. 
Pseudocode for this algorithm is given in Algorithm \ref{alg:frame_align}.

\newlength{\commentWidth}
\setlength{\commentWidth}{7cm}
\newcommand{\atcp}[1]{\tcp*[r]{\makebox[\commentWidth]{#1\hfill}}}

\begin{algorithm}[htb]
 \KwData{demo observation $O^D = (O^D_\text{RGB}, O^D_\text{Depth})$, \\
live observation $O^L=(O^L_\text{RGB}, O^L_\text{Depth})$, and\\
demo segmentation $S^D$.}
 \KwResult{transformation $T_{R,t}$}
 {
 \setstretch{1.15}
  \tcp{Compute flow from RGB}
 $F = \text{Flow}(O^D_\text{RGB},O^L_\text{RGB})$\; 
 \tcp{Compute live segmentation}
 $S^L = \text{warp}(S^D, F)$\;
\tcp{Apply mask to depth image}
 $O'^{D}_\text{Depth},O'^L_\text{Depth} = O^D_\text{Depth}[S^D],O^{L}_\text{Depth}[S^L]$\;
 \tcp{Unproject to 3D pointcloud}
 $p^D,p^L = \text{Unproj}(O'^D_\text{Depth}, O'^L_\text{Depth})$\;
 \tcp{Fit transformation T}
 $T_{R,t} = \adjustlimits \argminB_{R \in SO(3), t \in R^3} \Sigma (R \cdot p^D +t) - p^L$\; 
 }
 \vspace{.5em}
 \caption{\textbf{FrameAlign}: Procedure to find the transformation T that aligns the current frame with a demonstration frame for the foreground region defined by $S^D$. Figure \ref{fig:method} depicts this computation.}
 \label{alg:frame_align}
\end{algorithm}

\textbf{Sequence Tracking:}
In order to imitate a complete task, such as grasping an object, we need to successively align with respect to a sequence of demonstration images. To this end, when the live image is sufficiently close to the target image, as defined by a threshold, we step over to the next target image from the demonstration. As converging to each correct alignment takes some time, the tracking process can be accelerated by sampling only every n$^\text{th}$ demonstration frame. Moving the attention from one object to another requires a new target frame with the foreground mask on the new object. 

We need not generalize over some control quantities such as gripper state; for these we just copy the demonstration actions from the corresponding trajectory step. 
Interestingly, as long as our method correctly aligns all other dimensions an orthogonal dimension can be copied from the demonstrations. This allows aligning the in-plane components of the relative orientation and copying the height of the gripper.


To account for the time it takes the gripper to close, we delay progressing to the next frame accordingly. 
Pseudocode for the sequence tracking is given in Algorithm \ref{alg:sequence_track}.

\begin{algorithm}[htb]
 \KwData{demo. observations and segmentations $O^{D}, S^{D}$ \\ live observations $O^L$, demo length $N$, and distance threshold $\delta$. }
 \KwResult{imitated interaction }
 \setstretch{1.1}
 $i = 0$\tcp*{init. demo. frame index}
 $T = \infty$\tcp*{init. rel. transformation}
 \While{$i < N$ }{
     \While{$|T| > \delta$}{
         $T = \text{FrameAlign}(O_i^D, S_i^D, O^L)$\tcp*{get T}
         $a = \text{Action}(O^D, T)$\tcp*{get action}
         $O^L = \text{RobotAct}(a)$
         
     }
     $i=i+1$\tcp*{increase demo. frame}
     $T = \infty$\;
 }
 \caption{\textbf{Sequence Tracking}: Continuously align, and when close enough continue to next demonstration frame.}
 \label{alg:sequence_track}
\end{algorithm}

\textbf{Task Modularity:} Our method can combine individual subtasks into a multi-step task. This is done by switching the object segmented as foreground object during the demonstration. An example is shown in Figure \ref{fig:tasks}, where the first subtask is to grasp the wheel (the foreground mask is on the wheel) before the focus switches to the screw to connect the wheel with the screw. 

\section{Experiments}

We tested the end-to-end performance of our system on four manipulation tasks and in a localization experiment that is inspired by a real world industrial use case. 
Since camera pose estimation plays a key role in our setup, we additionally tested this component in isolation and quantified its performance relative to alternative pose estimation methods. Finally we tested the robustness of our system to variations in scene geometry and appearance.

\textbf{Setup:} The experimental setup consists of a KUKA iiwa arm with a WSG-50 two finger parallel gripper and an Intel SR300 projected light RGB-D camera mounted on the flange for an eye-in-hand view.
The camera is attached to a 3D-printed mount and faces towards the point between the fingertips.
The setup, which is shown in Fig. \ref{fig:tasks}, allows us to record depth images, which we use in our geometric fitting procedure.
The movement of the end effector is restricted such that the gripper always faces downwards;
it is parameterized as a 5 DoF continuous action $a = [\Delta x ,\Delta y, \Delta z,\Delta \theta,a_{\text{gripper}}]$ in the end effector frame.
$ \Delta x, \Delta y, \Delta z$ specify a Cartesian offset for the desired end effector position, $\Delta \theta$ defines the yaw rotation of the end effector, and $a_{\text{gripper}}$ is the gripper action that is mapped to the binary command to open or close the fingers. 

Our state observation consists of 640$\times$480 pixel RGB-D camera images and proprioceptive state vector consisting of the gripper height above the table, the angle that specifies the rotation of the gripper, and the width of the gripper fingers.
The optical flow is computed at the same resolution.

\subsection{Manipulation Experiments} 
Our approach is demonstrated by experiments on four example tasks: grasping a wooden block, inserting a block into a shape sorter, and grasping and inserting a wheel onto a screw. 
Solving these tasks requires precise movements; for example, an error of 4 mm is enough to make the insertion tasks fail. During the experiments we tested the robustness of our method by varying the object positions. To test the reactiveness of the approach we also moved the objects during task execution and varied the lighting conditions. Examples of this are shown in the supplemental videos.

\begin{figure}[!htb]
\centering
\begin{tabular}{c@{ }c}
\includegraphics[width=.48\linewidth]{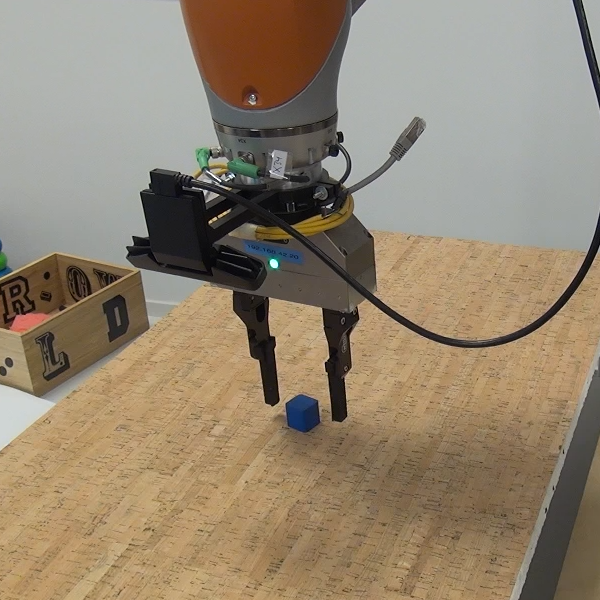} & \includegraphics[width=.48\linewidth]{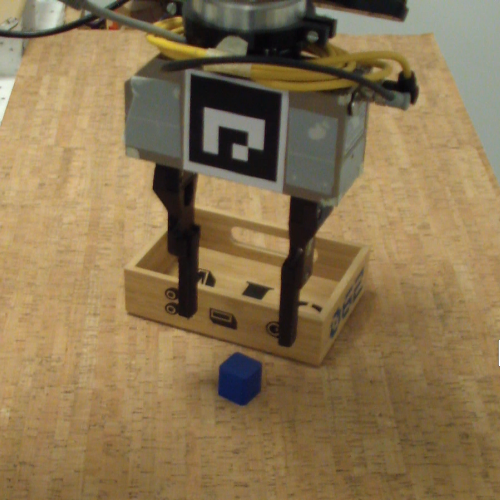} \\
  (a) Grasping Task & (b) Pick-and-Stow Task \\
    \\
 \includegraphics[width=.48\linewidth]{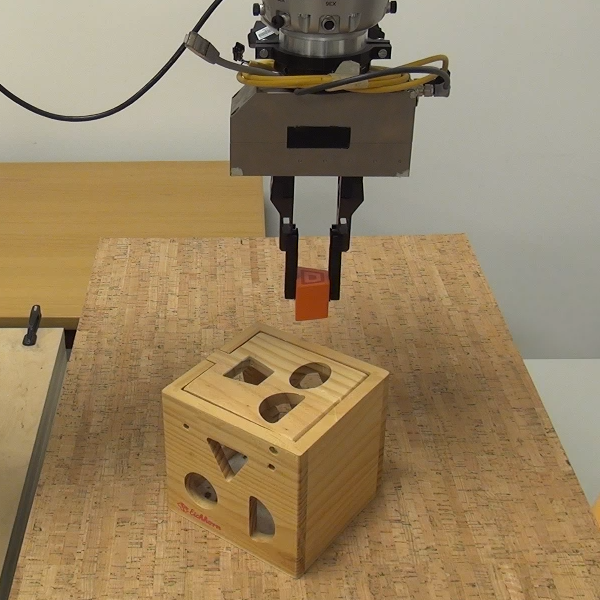} &
 \includegraphics[width=.48\linewidth]{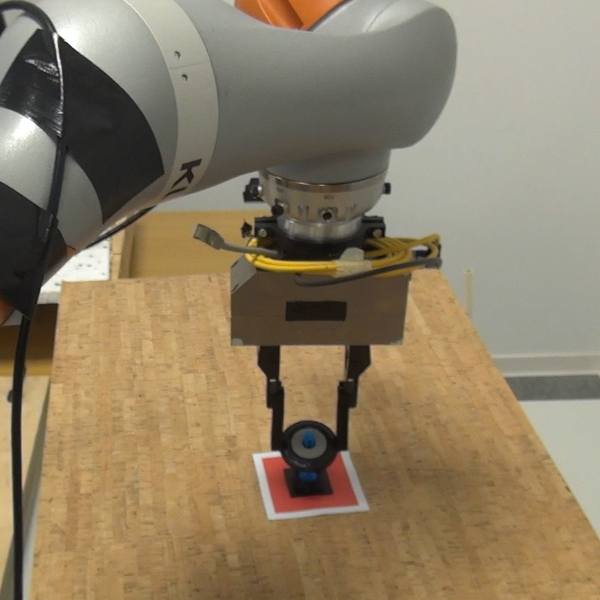} \\
 (c) Shape-Sorter Task & (d) Wheel Task
 \end{tabular}
 \caption{Images of manipulation tasks being completed, shown from an external perspective.}
 \label{fig:tasks}

\end{figure}

\textit{Grasping Blocks:}
A small, 25 mm wide, wooden block must be grasped and lifted from a table surface. 
This task is shown in Fig. \ref{fig:tasks} (a).

\textit{Pick-and-Stow:}
A small wooden block needs to be grasped and lifted from a table surface to be dropped into a box.

\textit{Shape-Sorter:}
A block must be inserted into a shape-sorting cube. 
This requires precise positioning due to the tight fit of the opening to the block.
We start with the block already grasped.
This task is shown in Fig. \ref{fig:tasks} (c).

\textit{Wheel Insertion:} This task uses parts from a toy construction set. 
A wheel must be grasped and inserted onto a screw that is held in a vertical position. This task is shown in Fig. \ref{fig:tasks} (d).


\textbf{Results:} The success rates for our manipulation experiments are shown in Table \ref{tab:success_rates}, examples are also shown in the supplemental video.
During testing, we used the same position distributions for task between methods as the success rate depends on the variation in the environment. 
Our method achieves high success rates. In addition, we outperform ACGD~\cite{Hermann19}, a recent approach that uses demonstrations to generate a curriculum for reinforcement learning. In contrast to ACGD, FlowControl does not require a simulation environment for task learning, which is difficult and time consuming to set up. 

\begin{table}[ht]
\begin{center}
\caption{Success rates for different manipulations tasks.}
\begin{tabular}{l l | c }
 Task &Method & Success Rate  \\ 
 \hline
 Pick-Stow & ACGD \cite{Hermann19} & 17 / 20  \\ 
 Pick-Stow & (Ours) & 19 / 20  \\
 Shape-Sorter & (Ours)  & ~8 / 10  \\
 Wheel Insertion& (Ours)  & ~9 / 10  \\
 
\end{tabular}
\vspace{1em}

\label{tab:success_rates}
\end{center}
\end{table}

Despite the generally good performance, we also identified failure cases.
A predictable source of problems were occlusions. These occurred, for example, in the pick-and-stow task when the box occluded the cube. 

Starting too far away from a target frame of the demonstration also leads to failure. FlowNet2 works within a given range of displacements. If the target is too far away, the optical flow will not find the correspondence anymore. Especially large rotations are a problem for optical flow. Section \ref{sec:fitting_experiments} quantifies the robustness of FlowControl in greater detail.
As the flow algorithm works for limited displacements in image space, starting demonstrations with the robot further away from the objects allows for coping with bigger displacements, as these appear smaller.

Optical flow is also more likely to fail when 
confusing background flows are present.
This is not prevented by the foreground mask as the flow algorithm receives as input the whole image and information from this may confound the flow computation. When running the controller with high velocities, a single wrong optical flow estimate can move the robot out of the convergent zone.
Running the controller at smaller velocities allows such errors to be corrected, reducing the chance of failure on the task.  

Other practical examples of failure were due to low illumination combined with fixed exposure times and grasping attempts snapping the object out of the visual field.

\subsection{Navigation Experiment}

\begin{figure}[!htb]
\centering
 \includegraphics[width=.60\linewidth]{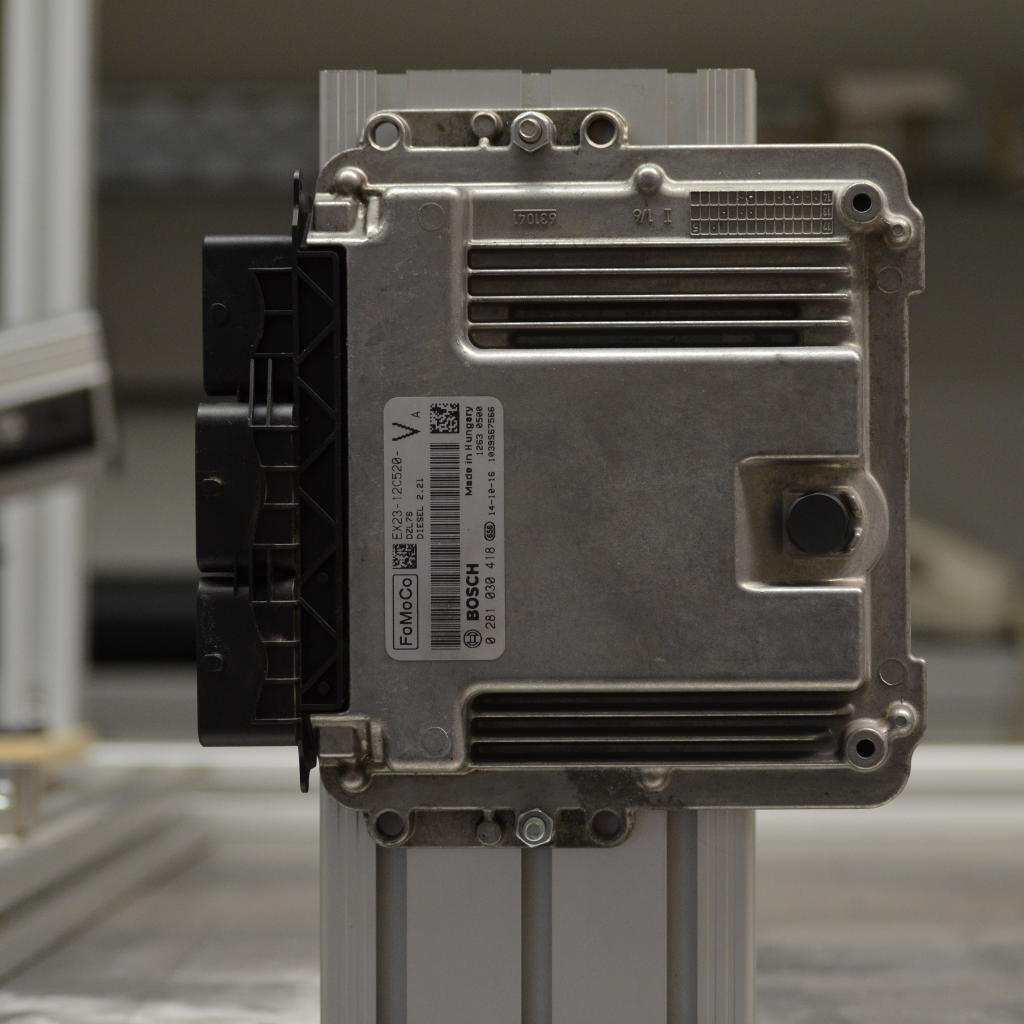} 
 \caption{Navigation task in an industrial setting showing an Engine Control Unit, from the robot's perspective, with respect to which a reference position has to be reached in order to fasten bolts.}
 \label{fig:nav_task}
\end{figure}

To demonstrate that the proposed method is directly applicable in an industrial setting, we evaluated a practical localization task. 
This task is based on an automotive assembly scenario in which a nut-runner must fasten nuts to fix an Engine Control Unit (ECU) into position.
The nut-runner must be positioned precisely in order to engage the bolts. Our method can visually align the end-effector resulting in greater robustness to variation in workpiece placement. The task is shown in Fig. \ref{fig:nav_task}.

We randomized the initial positions of the end-effector in the camera plane and measured how precisely it can return to the given reference position according to the robot's state-estimation. 
This resulted in a precision of $\pm 1$ mm, also measured using the robot's state estimation system. 
This test was repeated five times, with starting positions of up to 8 cm from the target position.
In these experiments, our reference image was taken with different lighting.

\subsection{Fitting Experiment}
\label{sec:fitting_experiments}
We evaluated the pose estimation component of our system using a proxy task with pre-recorded images.
Instead of moving an object with respect to the camera, we moved the camera with respect to a static object and recorded multiple views.
Visual markers were added to the scene to determine the relative camera pose between views; these were calculated using the FreiCalib tool \cite{Zimmermann20}.
The pose estimation algorithms must estimate this relative pose.
We evaluate several baselines. The simplest is a zero pose change prediction. The SIFT baseline is evaluated similar to \cite{Skrypnyk04}. We also compare to DeepTAM \cite{Zhou18}, a learned algorithm that estimates depth and relative pose given to images.

In this setup, the static background could be used to infer the relative pose. 
To mitigate this, we again masked our computed features with the demonstration segmentation, except for DeepTAM, where this was not possible, because it is a monolithic system that produces a pose.
As both SIFT matching and optical flow methods have outliers, we substitute zero pose change predictions for rotations larger than $\frac{\pi}{4}~\text{rad}$, and translations larger than $250~\text{mm}$.

\begin{figure}[!htb]
\centering
\begin{tabular}{c@{ }c}
 \includegraphics[width=.48\linewidth]{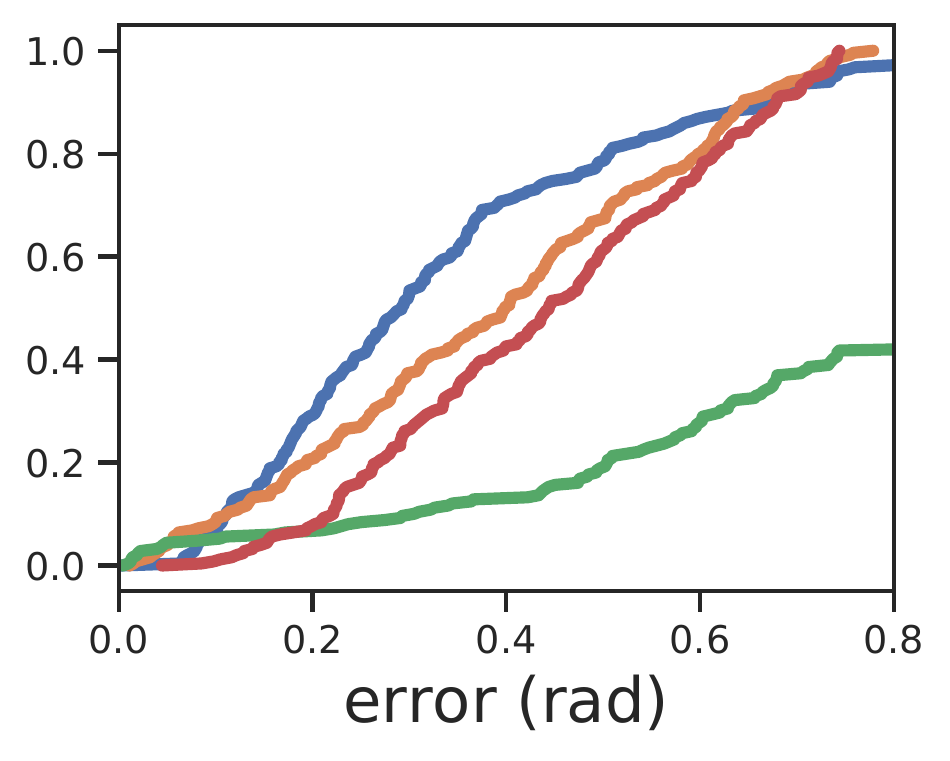} &
 \includegraphics[width=.48\linewidth]{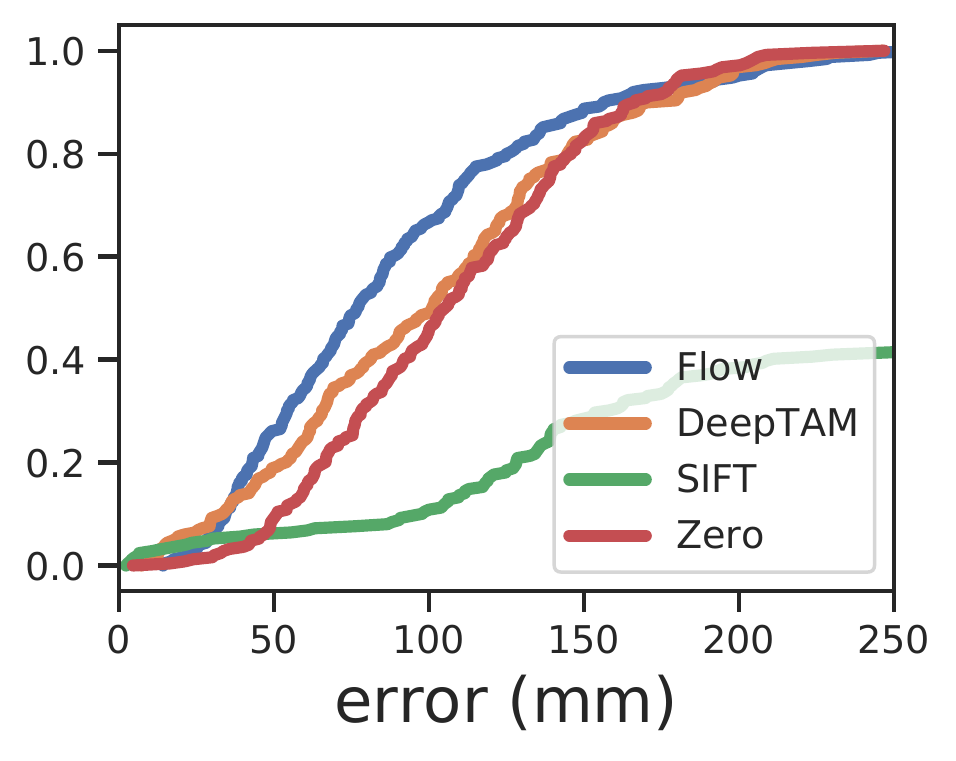} \\
 Relative Orientation Error & Relative Translation Error \\
\end{tabular}
 \caption{Comparison of the proposed method to baselines showing percent of samples (y-axis) below a given error threshold (x-axis).}
 \label{fig:rot_trans_err}
\end{figure}

\begin{figure}[!htb]
 \centering
 \includegraphics[width=.95\linewidth]{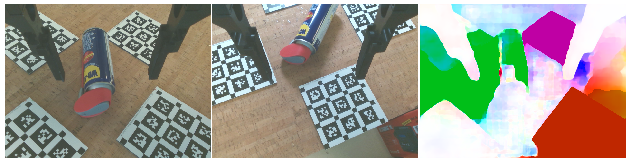}
 \caption{Example of erroneous optical flow due to large rotation and limited frame overlap. Examples of good flow are shown in Fig. \ref{fig:teaser} and Fig. \ref{fig:method}. White indicates zero flow magnitude.}
 \label{fig:bad_flow}
\end{figure}

This evaluation was done on a selection of 250 views with a relative rotation of less than $\frac{\pi}{4}~ \text{rad}$.
The results are shown in Figure \ref{fig:rot_trans_err}.
SIFT based relative pose detection performed badly, completely failing for most views.
This resulted in worse performance than the zero pose change baseline.
DeepTAM performs slightly better than the zero rotation. However, as this approach is designed for smaller pose differences it often underestimates changes.
Our flow-based approach performs best for most samples, although it still has outliers in cases where the flow computation failed. An example of this is shown in Figure \ref{fig:bad_flow}. This usually occurs for a combination of large displacements and rotations.

\subsection{Generalization Experiments}

In contrast to classical fixed visual servoing approaches, FlowNet has been trained to be invariant to miscellaneous effects, such as lighting changes and partial occlusion.
This helps it find correspondences even when objects in the demonstration do not match exactly.
For simplicity, we limited the generalization experiments to the grasping task. 
We recorded a demonstration with one object and then tested if this demonstration generalizes to objects of different shapes and sizes. Examples of this are shown in Figure \ref{fig:generalization} and in the supplemental video.
FlowControl is able to cope with variation in both color and shape.

\begin{figure}[!htb]
\centering
 \includegraphics[width=1\linewidth]{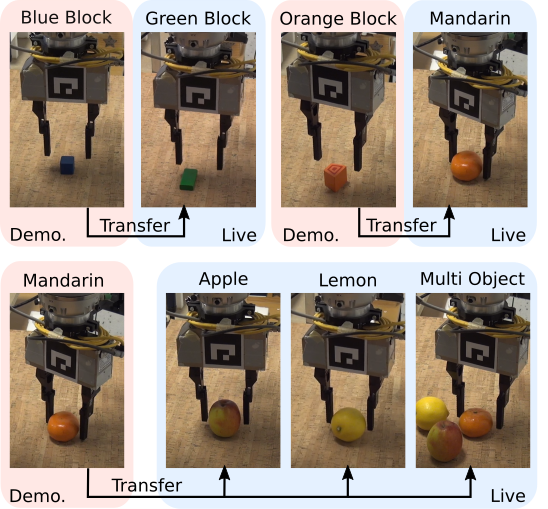}
 \caption{ Generalization experiments showing transfer of grasping with objects that are different from the object used for demonstration. 
 }
 \label{fig:generalization}
\end{figure}




\section{Discussion and Conclusion}
We presented a practical, data-efficient method for visual servoing from optical flow. Our method works with single demonstrations and is able to handle significant variations in the geometric arrangement as well as visual appearance of the task. We demonstrated the effectiveness of our method on a series of robotic manipulation experiments. In addition, we provided a quantitative assessment of the pose estimation part of our algorithm and combined this with a discussion of possible failure cases of our method. Finally, we also provide some experiments indicating that our method is able to generalize over substantial variation in geometry and appearance.

While FlowControl has many advantageous properties, it has natural limitations: it cannot yet do re-grasping and currently relies on manual segmentation to define the task.
Current failure cases include optical flow methods failing for large displacements. 
One could train optical flow specifically for the type of data distribution at hand: one with larger rotations and displacements, or for the specific objects that may appear in the task. 

Despite this, FlowControl satisfies an important aim; robotics algorithms should not merely solve one specific task, but instead obviate the need for task-specific engineering. With little manual effort, FlowControl solves a diverse set of tasks.


\bibliographystyle{IEEEtran}
\bibliography{references}

\end{document}